\documentclass[runningheads]{llncs}
\usepackage{graphicx}
\usepackage{multirow}
\usepackage{hyperref}
\usepackage{todonotes}

\usepackage{comment}
\begin{document}

\title{LLMs4OM: Matching Ontologies with \\ Large Language Models}

\titlerunning{Matching Ontologies with Large Language Models}
%
\author{
Hamed Babaei Giglou\orcidID{0000-0003-3758-1454} \and Jennifer D’Souza\orcidID{0000-0002-6616-9509} \and Felix Engel\orcidID{0000-0002-3060-7052} \and  S\"{o}ren Auer\orcidID{0000-0002-0698-2864}}
\authorrunning{Babaei Giglou et al.}
%
\institute{TIB Leibniz Information Centre for Science and Technology, Hannover, Germany\\
\email{\{hamed.babaei,jennifer.dsouza,felix.engel,auer\}@tib.eu}\\
}
%
%
%
\maketitle              
\begin{abstract}
Ontology Matching (OM), is a critical task in knowledge integration, where aligning heterogeneous ontologies facilitates data interoperability and knowledge sharing. Traditional OM systems often rely on expert knowledge or predictive models, with limited exploration of the potential of Large Language Models (LLMs). We present the LLMs4OM framework, a novel approach to evaluate the effectiveness of LLMs in OM tasks. This framework utilizes two modules for retrieval and matching, respectively, enhanced by zero-shot prompting across three ontology representations: concept, concept-parent, and concept-children. Through comprehensive evaluations using 20 OM datasets from various domains, we demonstrate that LLMs, under the LLMs4OM framework, can match and even surpass the performance of traditional OM systems, particularly in complex matching scenarios. Our results highlight the potential of LLMs to significantly contribute to the field of OM.
\end{abstract}

\section{Introduction} 
In the dynamic field of information and data management, ensuring the interoperability and integration of varied knowledge systems is critical. Ontologies play a key role in achieving semantic interoperability by providing a structured, understandable framework for both humans and machines~\cite{Stephan2007,ontology2001}. However, the proliferation of new ontologies presents challenges in aligning them for seamless communication across different systems~\cite{ZHANG20158,OSMAN202138}. Ontology matching (OM) emerges as a vital solution, automating the discovery of correspondences across ontologies~\cite{om2013}. The emergence of Large Language Models (LLMs) in natural language processing has revolutionized the traditional boundaries between human and machine understanding of language, making LLMs highly relevant for OM tasks. Despite initial efforts to apply LLMs to OM~\cite{OLaLa,ConversationalOM}, the rapid development of these models calls for an in-depth exploration of their potential in OM, which this study aims to provide, emphasizing the importance of OM and the promising capabilities of LLMs in addressing its challenges.

To pursue this objective, we present the LLMs4OM framework, which assesses diverse LLMs across various tracks and domains introduced within the Ontology Alignment Evaluation Initiative (\href{https://oaei.ontologymatching.org/}{OAEI})~\cite{OAEI23}. OM aims to map concepts between source $C_{source} \in O_{source}$ and target $C_{target} \in O_{target}$ ontologies. Formally, the task is to identify for any $C_s \in C_{source}$, possible $C_t \in C_{target}$ that $(C_s, C_t, S_{C_s \equiv C_t})$, where $S \in [0, 1]$ represents the likelihood of equivalence $C_s \equiv C_t$~\cite{OAEI11}.

An initial exploratory study using ChatGPT-4~\cite{ConversationalOM}, demonstrated the OM task via a conversational, naive approach, where ontologies $O_{source}$ and $O_{target}$ were fully inputted into the LLM to solicit matchings. This approach, however, highlighted two primary drawbacks: i) the limited context length LLMs can process, which may be exceeded by larger ontologies, and ii) the increased likelihood of erroneous or "hallucinated" responses due to the volume of information provided. To address these, LLMs4OM employs a dual-module strategy: first, using the Retrieval-Augmented Generation (RAG)~\cite{rag21} for candidate selection for a given query $C_{source}$ from a knowledge base of $C_{target}$, and then LLM-based matching, in a second module, for finer accuracy. This approach mitigates the limitations of direct LLM prompting by optimizing for the specific challenges of OM, demonstrating a strategic advancement in leveraging LLMs for ontology alignment. In our study using the LLMs4OM framework, we conduct extensive evaluations, beginning with the RAG module where we explore four retrieval methods: TFIDF~\cite{tfidf}, sentence-BERT~\cite{sbert}, SPECTER2~\cite{specter2}, and OpenAI text-embedding-ada~\cite{adaembedding}. Subsequently, within the LLM module, we pair these retrieval techniques with seven state-of-the-art LLMs: LLaMA-2~\cite{llama2}, GPT-3.5~\cite{chatgpt}, Mistral~\cite{mistral}, Vicuna~\cite{vicuna}, MPT~\cite{mpt}, Falcon~\cite{falcon}, and Mamba~\cite{mamba}, to assess their combined effectiveness. Furthermore, a detailed analysis based on our large-scale experiments framed three main research questions (\textbf{RQ}s). \textbf{RQ1}: What impact do the three concept representations (concept, concept-parent, concept-children), respectively have on improving matching efficacy? \textbf{RQ2}: For the RAG module, which retriever performs best per track? (\textit{RQ2.1}) Additionally, how does recall vary in the retrieval module across our different retrieval techniques employed? (\textit{RQ2.2}) \textbf{RQ3}: Which LLM performs best per track? (\textit{RQ3.1}) Furthermore, how does the performance of various LLMs differ across the three concept representations for the OM tracks? (\textit{RQ3.2}).

This study's empirical tests are varied not only in their approach but also in the range of ontological knowledge domains they cover. We evaluate LLMs across six tracks of the OAEI campaign, encompassing 20 datasets in total. The primary contributions of this paper are threefold: 1) Introduction of the LLMs4OM, an end-to-end framework that utilizes LLMs for OM; 2) A thorough empirical evaluation of seven state-of-the-art domain-independent LLMs and four retrieval models for their suitability to the various OM tasks; and 3) The source code implementation of the LLMs4OM framework released here \url{https://github.com/HamedBabaei/LLMs4OM}.
\vspace{-1mm}
\section{Related Work}
Ontology matching, a well-explored research area, has seen diverse methodologies, from traditional techniques~\cite{LogMap,AgreementMakerLight,LSMatch,ALIN,OAPT} to recent transformer-based methods~\cite{TEXTO,AMD,GraphMatcher,Matcha,PropMatch,Truveta,LaKERMap,SORBET,Matcha,BERTMap,ConversationalOM,OLaLa,he2023exploring}, each contributing to advancements in the field. Despite the proven effectiveness of conventional approaches, this work focuses on classifying ontology matching systems, especially those utilizing transformers~\cite{transformers}, into three categories based on their research goals: unsupervised learning, supervised learning, and LLM-based approaches.

Unsupervised learning methods in ontology matching often use embeddings for similarity assessments. Techniques such as TTEXTO~\cite{TEXTO}, PropMatch~\cite{PropMatch}, AMD~\cite{AMD}, and Matcha~\cite{Matcha} primarily leverage BERT~\cite{bert} variants (e.g., RoBERTa~\cite{roberta}, sentence-BERT) to generate ontology embeddings for these calculations. Additionally, some methods combine transformer models with multiple representations: TEXTO integrates GloVe~\cite{glove} with BERT, AMD pairs knowledge graph embeddings with BERT, GraphMatcher~\cite{GraphMatcher} combines universal sentence encoder~\cite{use} with graph learning techniques, and PropMatch uses sentence-BERT with TFIDF for enhanced matching accuracy. Supervised OM methods predominantly fine-tune transformer models. Truveta Mapper~\cite{Truveta} utilizes ByT5~\cite{byt5} on the Bio-ML track, employing a sequence-to-sequence approach. LaKERMap~\cite{LaKERMap} focuses on domain-specific tuning with Bio-ClinicalBERT~\cite{biocilinicalbert}. SORBETmatcher~\cite{SORBET} combines BERT with random walks and regression loss for ontology embeddings. Matcha-DL~\cite{Matcha} uses sentence-BERT in a semi-supervised setup with a dense network for candidate ranking. BERTMap~\cite{BERTMap} integrates unsupervised and semi-supervised strategies by initially fine-tuning BERT on ontology texts, and then refining mappings based on ontology structure.

Research on larger parameter models~\cite{ConversationalOM,he2023exploring,OLaLa} reveals significant strategies for ontology matching (OM). ~\cite{ConversationalOM} leverages prompt templates with LLMs to input source and target ontologies, showcasing OM potential. OLaLa~\cite{OLaLa} utilizes LLaMA-2 models and BERT retrievers to extract top-k matches from target ontologies for LLM prompts, refining final alignments with a precision matcher and filters. LLMap~\cite{he2023exploring} investigates Flan-T5~\cite{flant5} and GPT-3.5's zero-shot capabilities, focusing on concept labels and structural contexts.

\section{LLMs4OM -- Methodological Framework}
\begin{figure}
\includegraphics[width=\textwidth]{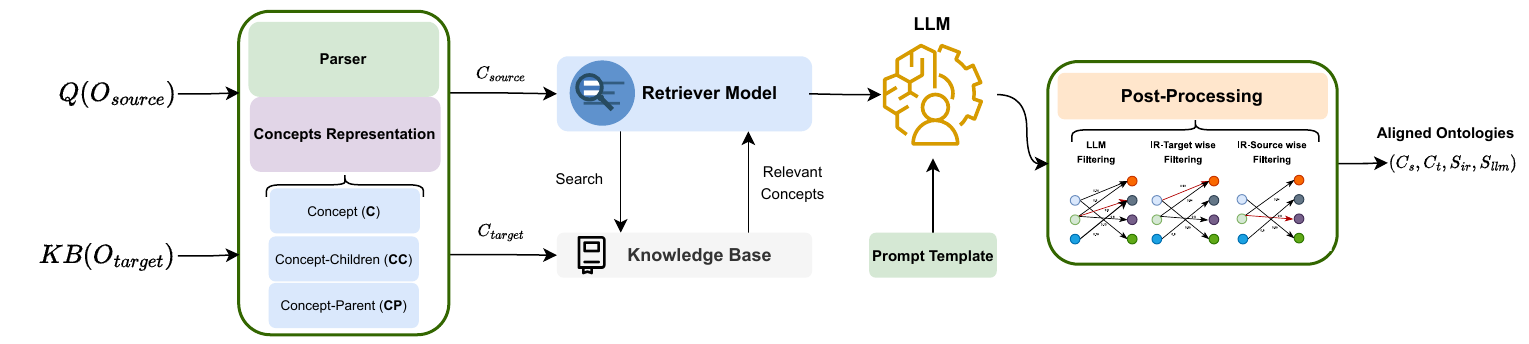}
\caption{Overview on LLMs4OM as an end-to-end framework for OM.} \label{llms4om}
\end{figure}

The LLMs4OM framework offers a RAG approach within various LLMs for OM. LLMs4OM uses $O_{source}$ as query $Q(O_{source})$ to retrieve possible matches for for any $C_s \in C_{source}$ from $C_{target} \in O_{target}$. Where, $C_{target}$ is stored in the knowledge base $KB(O_{target})$. Later, $C_{s}$ and obtained $C_t \in C_{target}$ are used to query the LLM to check whether the $(C_s, C_t)$ pair is a match. As shown in \autoref{llms4om}, the framework comprises four main steps: 1) Concept representation, 2) Retriever model, 3) LLM, and 4) Post-processing.

\noindent\textbf{1) Concept representation.} Within this module, we process the ontologies, to extract the child, parent, and concept-specific representations of ontology elements. These representations will be utilized to generate three distinct input representations: i) Concept ($C$), a foundational representation that encapsulates the core characteristics of a standalone concept within the ontology, ii) Concept-Parent ($CP$), extending beyond individual concepts, this representation establishes the hierarchical relationships by incorporating information about the parent concepts, and iii) Concept-Children ($CC$) complementing the $CP$ representation which focuses on the descendants of a given concept. These variant representations ensure a comprehensive understanding of ontologies, capturing both individual concepts and their hierarchical relationships, thus supporting the complete nature of ontologies. Subsequently, minor preprocessing is carried out to acquire clean textual data by converting representations to lowercase and removing punctuation. Finally, the parents in $CP$ and children in $CC$ are merged to create a list of parents and children associated with concepts, respectively.

\noindent\textbf{2) Retriever model}. First, an embedding extractor model operates by extracting embeddings for $C_{target} \in O_{target}$ and forming an embedding knowledge base for all $C_{target}$. Next, for a given $C_{s} \in C_{source}$, using the embedding extractor model, a $C_{s}$ embedding is generated to calculate cosine similarity across all $C_{target}$, to identify $top_k$ most similar candidates for alignments.  The retrieval model will result in a $\{(C_s, C_{t_1}), ..., (C_s, C_{t_k})\}$ pairs with similarity score of $S_{ir}$ per pair. For other input representations, $C$ can be substituted with $CC$ to include children or with $CP$ to integrate the parent of $C$ within the representations for the retrieval model.

\noindent\textbf{3) \textsc{LLM}.} Using obtained $\{(C_s, C_{t_1}), ..., (C_s, C_{t_k})\}$ pairs from the retrieval model, each pair is verbalized as text and replaced in the prompt template to input LLMs. Subsequently, employing the LLM prompting technique~\cite{liu2021pretrain}, inputs are categorized into "yes" and "no" classes using label words such as yes/true/right for the "yes" class and no/false/wrong for the "no" class. Later, the confidence score of $S_{llm}$ is derived from the probabilities assigned to the "yes" and "no" classes label words corresponding to the obtained pairs. The following prompt template is designed to use $C$, $CC$, or $CP$ representation of ontology concepts.
\begin{quote}
\small
    \textit{Classify if two concepts refer to the same real-world entity or not (answer only yes or no).\textbackslash{}n\#\#\# First concept:\textbackslash{}n\{$C_{s}$\}\textbackslash{}n[Parents$|$Children]: \{$CP|CC$\}\textbackslash{}n\#\#\# Second concept:\textbackslash{}n\{$C_{t}$\}\textbackslash{}n [Parents$|$Children]: \{$CP|CC$\}\textbackslash{}n\#\#\# Answer:}
\end{quote}
Where in the template \{$C_{s}$\} and \{$C_{t}$\} are placeholders for pair concepts. The notation \textit{\small{"[Parents$|$Children]: \{$CP|CC$\}"}} offers flexibility in representing ontology concepts, allowing for the inclusion of either parent or children concepts via $CP$ and $CC$ representations.

\noindent\textbf{4) Post-processing.} After obtaining the retrieval model similarity score of $S_{ir}$ and LLM's confidence scores of $S_{llm}$ for "yes" and "no" classes, we conducted hybrid post-processing to obtain final pairs that match among ${(C_s, C_{t_1}), ..., (C_s, C_{t_k}))}$. The hybrid post-processing involves three steps:
\begin{enumerate}
    \item \textit{Confidence-driven filtering by LLM}: First, predicted pairs with the "no" class are disregarded. Then, pairs with $S_{llm} > 0.7$ for the "yes" class are retained.
    \item \textit{The high precision matcher}: This step applies to the retrieval model similarity score using $S_{ir} > 0.9$. The resulting output consists of exact matches.
    \item \textit{Cardinality-based filtering}: Implemented to prevent multiple matches per $C_{source}$ or $C_{target}$ concepts. 
\end{enumerate}
This yields $(C_s, C_t, S_{ir}, S_{llm})$ as the set matching between concepts.

\section{LLMs4OM -- Ontology Matching Evaluations}

\subsubsection{Evaluation Datasets -- OAEI Tracks \& Tasks.}
We selected five tracks from the OAEI campaign, covering diverse domains, and utilized three setups, i.e. concept, concept-children, and concept-parent, for our experiment. These configurations aim to identify the most effective ontology representation for OM, particularly focusing on the equivalence matching problem. The chosen tracks includes: \textsc{Anatomy}) Anatomy~\cite{anatoym} (Mouse-Human),  \textsc{biodiv}) Biodiversity and Ecology~\cite{biodiversity} (8 tasks), \textsc{Phenotype}) Disease and Phenotype~\cite{phenotype} (DOID-ORDO and HP-MP), \textsc{CommonKG}) Common Knowledge Graphs~\cite{commonkg} (Nell-DBpedia and YAGO-Wikidata), \textsc{Bio-ML}) Biomedical Machine Learning~\cite{bioml} (5 tasks), and \textsc{MSE}) Material Sciences and Engineering~\cite{mse} (MI-EMMO and MI-MatOnto) OAEI tracks which resulted in 20 tasks/datasets. 

\subsubsection{Evaluation Models -- Retrievers \& LLMs.}
As already introduced earlier, in this work, this study evaluates 7 state-of-the-art LLMs across 4 retriever models using the LLMs4OM framework. We assess retrieval models including TFIDF~\cite{tfidf}, sentence-BERT~\cite{sbert}, SPECTER2~\cite{specter2}, and OpenAI text-embedding-ada~\cite{adaembedding}. Afterward, we combine these with LLMs (number of parameters written in parenthesis) such as LLaMA-2 (7B)~\cite{llama2}, GPT-3.5 (174B)~\cite{chatgpt}, Mistral (7B)~\cite{mistral}, Vicuna (7B)~\cite{vicuna}, MPT (7B)~\cite{mpt}, Falcon (7B)~\cite{falcon}, and Mamba (2.8B)~\cite{mamba} to measure their effectiveness for OM.

\subsubsection{LLMs4OM Results.} 
\begin{figure}[h]
\includegraphics[width=\textwidth]{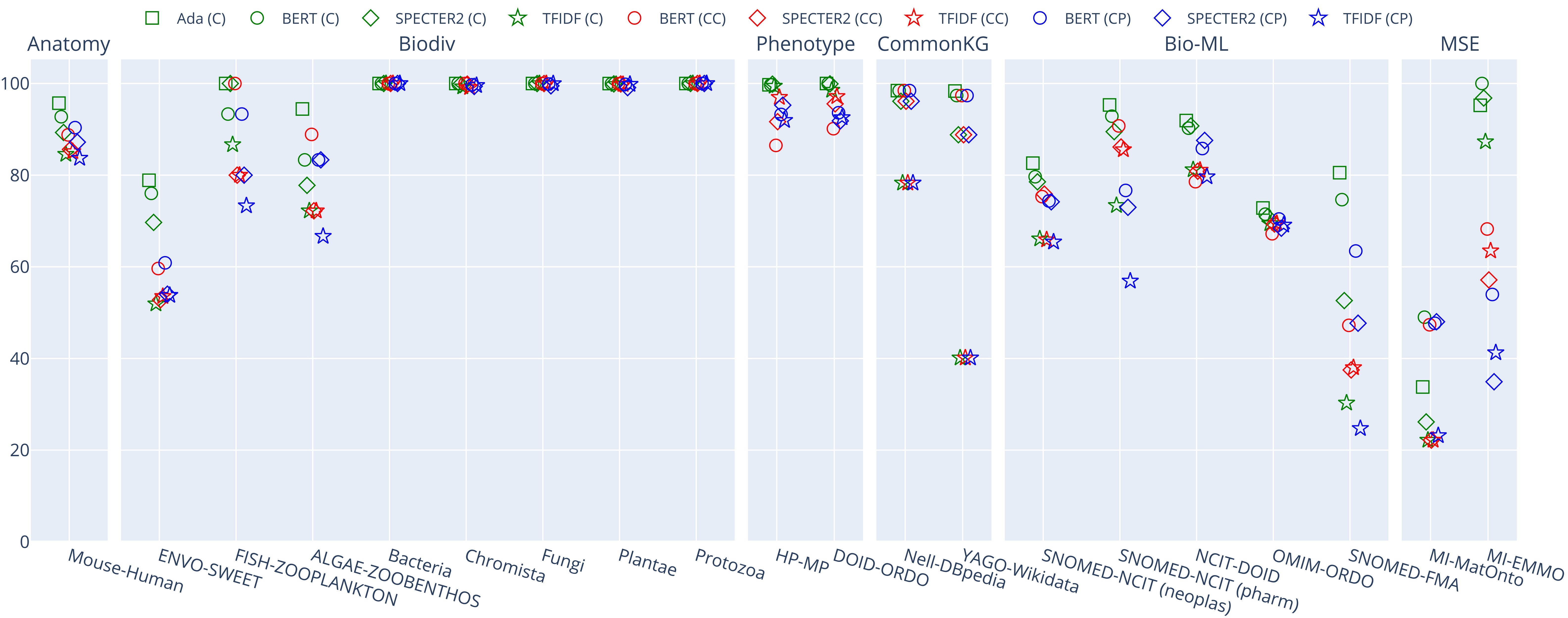}
\caption{Comparing retrieval models using recall and $top_k=5$.}\label{irresults}
\vspace{-4mm}
\end{figure}

For each track, retriever models with $top_k=5$ are evaluated across proposed concept representations and results are reported in~\autoref{irresults}. The assessment includes 7 LLMs with $C$, $CC$, and $CP$ input representations, along with retrievers like \textit{text-embedding-ada} and \textit{sentence-BERT}, detailed in~\autoref{llm_ir_results}. Approximately 50 runs per dataset were conducted, providing foundational results for further analysis (the complete results are indicated in supplementary material). We focus on zero-shot evaluations of LLMs and retrieval models in addressing our research questions.
\begin{table}[h]
    \centering
    \small
    \caption{Best zero-shot performer model results across 20 tasks, using 7 LLMs, 3 concept representations (C, CP, CC), and 2 retriever models. Bold denotes where LLM4OM using respective models outperforms OAEI 2023 OM systems. The "OAEI" column displays the top F1-score from OAEI 2023.}
    \label{llm_ir_results}
    \begin{tabular}{|l|l|c|c|c|c|c|}
        \hline
        \textbf{Track} & \textbf{Tasks} &\textbf{Prec} &\textbf{Rec} & \textbf{F1} &\textbf{Best Model} &\textbf{OAEI}\\
        \hline
        \textit{\textsc{Anatomy}} & Mouse-Human &  90.82 &  87.46 & 89.11 & GPT-3.5(C)+Ada&94.10\\
        \hline
         \multirow{8}{*}{\textit{\textsc{Biodiv}}} &  ENVO-SWEET &  59.00	& 51.67 & 55.09& Mistral(C)+Ada&71.40\\
         &  FISH-ZOOPLANKTON & 100 &  80.00 & 88.88 & LLaMA-2(C)+Ada&92.80 \\
         & ALGAE-ZOOBENTHOS   & 100 & 38.88 & \textbf{56.00} & Mistral(C)+Ada&50.00\\
         & TAXR-NCBI(Bacteria) & 67.96 &  99.42 & \textbf{80.74} & GPT-3.5(CP)+Ada& 74.80\\
         & TAXR-NCBI(Chromista) & 69.87 & 98.07 & \textbf{81.61} &  GPT-3.5(CP)+Ada&77.30\\
         & TAXR-NCBI(Fungi) &  86.97 & 99.08 & \textbf{99.63} & GPT-3.5(CP)+Ada&89.10\\
         & TAXR-NCBI(Plantae) & 82.59 & 96.34 & \textbf{88.94} & GPT-3.5(CP)+Ada&86.60 \\
         & TAXR-NCBI(Protozoa) & 86.06 & 98.59 & \textbf{91.90} & GPT-3.5(CP)+Ada&85.70\\
         \hline
         \multirow{2}{*}{\textit{\textsc{Phenotype}}} & DOID-ORDO & 85.79 & 94.26 & \textbf{89.83} & Mistral(CP)+BERT&75.50\\
         & HP-MP &  76.67 & 95.40 & \textbf{85.01} & Mistral(CP)+BERT&81.80\\
         \hline
         \multirow{2}{*}{\textit{\textsc{CommonKG}}} & Nell-DBpedia & 100  & 89.14 & 94.26 & GPT-3.5(C)+Ada&96.00  \\
         & YAGO-Wikidata& 100 &  85.52 & 92.19 & LLaMA-2(C)+Ada&94.00\\
         \hline
         \multirow{5}{*}{\textit{\textsc{Bio-ML}}} & NCIT-DOID (disease)& 86.19 &  80.06 & 83.01 & GPT-3.5(C)+Ada&90.80 \\
         & OMIM-ORDO (disease) & 71.80 & 57.96 & 64.14 & GPT-3.5(CC)+Ada&71.50\\
         & SNOMED-FMA(body) & 21.12 & 32.60 & 25.64 &  GPT-3.5(CP)+Ada&78.50\\
         & SNOMED-NCIT(neoplas)&  46.96 & 52.96  & 49.47 & GPT-3.5(CP)+Ada&77.10 \\
         & SNOMED-NCIT(pharm)&  81.84 & 58.19 & 68.02 & GPT-3.5(CC)+Ada&75.20  \\
         \hline
         \multirow{2}{*}{\textit{\textsc{MSE}}}& MI-EMMO & 96.66 & 92.06 & \textbf{94.30} & LLaMA-2(CC)+BERT&91.80 \\
         & MI-MatOnto & 89.70 & 20.19 & 32.97 & MPT(C)+BERT&33.90\\
        \hline
    \end{tabular}
    \vspace{-3mm}
\end{table}

\noindent{\textbf{RQ1: What impact do the three concept representations, respectively have on improving matching efficacy?}} We address this question by analyzing the findings presented in~\autoref{irresults}, demonstrating the superiority of the $C$ representation across all 20 tasks for retrieval models using the proposed method. Additionally, in~\autoref{llm_ir_results}, we find $C$ excelling in 6 tasks, while $CP$ outperforms in 9 tasks. Notably, based on observation of results in~\autoref{irresults}, SNOMED-FMA tasks from \textsc{Bio-ML} show high sensitivity to input representation. Furthermore, the inclusion of information about concepts, i.e. parents or children, shifts representations towards other concepts within the ontology. However, such information proves to be valuable for LLMs in enhancing their understanding of concepts, as evidenced in the results.

\noindent{\textbf{RQ2.1. [Retrieval module] Which retriever performs best per track?}} Given the results in~\autoref{irresults}, we analyze this question. Across tracks like \textsc{Anatomy}, \textsc{Biodiv}, \textsc{Phenotype}, \textsc{CommonKG}, and \textsc{Bio-ML}, OpenAI \textit{text-embedding-ada} consistently outperforms. However, in \textsc{MSE} track tasks, \textit{sentence-BERT} emerges as the standout performer. Specifically, for the challenging MI-MatOnto task, \textit{sentence-BERT} achieves a 49.00\% recall. Combining top retrievers, \textit{text-embedding-ada} and \textit{sentence-BERT}, with LLMs, as shown in~\autoref{llm_ir_results}, highlights \textit{sentence-BERT}'s suitability for \textsc{Phenotype} and \textsc{MSE} tracks, while \textit{text-embedding-ada} excels in the remaining 4 tracks. These findings underscore the importance of selecting appropriate retrievers tailored to specific task requirements in LLMs4OM.

\noindent{\textbf{RQ2.2: [Retrieval module] How does recall vary in the retrieval module across our different retrieval techniques employed?}} We investigate this question by comparing retriever models across different values of $top_k \in [5, 10, 20]$. On average, for $top_k=5$, the retrieval models achieve a recall of 82.09\%, increasing to 84.66\% for $top_k=10$, and further to 86.82\% for $top_k=20$. Specifically, given the results in~\autoref{irresults}, when considering $top_k=5$, the \textit{text-embedding-ada} retriever achieves a recall of 90.88\%, followed by \textit{sentence-BERT} with 86.09\%, \textit{SPECTER2} with 82.10\%, and \textit{TFIDF} with 75.15\%, highlighting the superior performance of \textit{text-embedding-ada} and \textit{sentence-BERT}. However, it's important to note that higher values of $top_k$ lead to increased time complexity and longer waiting times. Consequently, we may opt to sacrifice a 4\% average recall with $top_k=20$ in exchange for reduced waiting times using $top_k=5$.

\noindent{\textbf{RQ3.1: [LLM module] Which LLM performs best per track?}} We examine this question given the results in~\autoref{llm_ir_results}. The \textbf{Best Model} column in the table showcases the top-performing models, starting with GPT-3.5, followed by Mistral-7B, LLaMA-2-7B, and finally MPT-7B among the 7 LLMs. The summary of best model results concerning OM systems proposed in OAEI 2023~\cite{OAEI23} using F1-score are as follows: for MI-EMMO LLaMA-2-7B 94.30\% $>$ Matcha 91.8\% \cite{Matcha}, for HP-MP Mistral-7B 85.01\% $>$ LogMap 81.8\%~\cite{LogMap}, for DOID-ORDO Mistral-7B 89.93\% $>$ AML 75.5\%~\cite{AgreementMakerLight}, for ALGAE-ZOOBENTHOS Mistral-7B 56.00\% $>$ OLaLa 50.0\%~\cite{OLaLa}, for TAXR-NCBI(Bacteria) GPT-3.5 80.74\% $>$ LogMapLt 77.3\%~\cite{LogMap}, for TAXR-NCBI(Fungi) GPT-3.5 99.63\% $>$ OLaLa 89.9\%~\cite{OLaLa}, for TAXR-NCBI(Plantae) GPT-3.5 88.94\% $>$ OLaLa 86.6\%~\cite{OLaLa}, and for TAXR-NCBI(Protozoa) GPT-3.5 91.90\% $>$ OLaLa 85.7\%~\cite{OLaLa}.

\noindent{\textbf{RQ3.2: [LLM module] How does the performance of various LLMs differ across the three concept representations for the OM tracks?}} Using results from~\autoref{llm_ir_results}, we find LLMs perform better with additional contexts like parents or children, as seen in tasks across \textsc{Biodiv}, \textsc{Phenotype}, and \textsc{Bio-ML} tracks. In \textsc{Biodiv}, $CP$ consistently boosts LLM performance, especially in TAXR-NCBI tasks. Similarly, \textsc{Phenotype} tasks show improved results with $CP$ representations, notably in DOID-ORDO (89.83\%) and HP-MP (85.01\%). However, \textsc{Bio-ML} tasks exhibit mixed outcomes; some like NCIT-DOID perform well (83.01\%) with $C$ representation, while others like SNOMED-FMA (25.64\%) struggle even with $CP$ representation. In \textsc{MSE}, tasks vary greatly; for example, MI-EMMO achieves 94.30\% success with LLaMA-2-7B and $CC$ representation. This highlights the importance of selecting the right model architecture and contextual representation for each task. Overall, this analysis stresses the significance of context in LLMs across diverse domains, emphasizing the need for tailored approaches based on task specifics.

\section{Discussion}
\textbf{Benefits of using our RAG technique for OM.} Integrating retrieval with LLMs yields benefits. Initially, querying LLM with all pairs led to impractical $O(n^2)$ time complexity, particularly with larger datasets. However, integrating retrieval reduces complexity to linear $O(kn)$, enabling faster processing while preserving LLM-generated confidence scores. Additionally, providing all ontologies at once to the model, as seen in~\cite{ConversationalOM}, results in mixed outputs, posing challenges in computing matching scores and increasing the risk of high hallucination, especially with larger ontologies.


\noindent\textbf{Low performance on the \textsc{Bio-ML} track.} LLMs4OM showed low performance compared to traditional methods across the \textsc{Bio-ML} track tasks. Analyzing their performance with two retrievers, we found an average F1-score of 53\% with \textit{text-embedding-ada} and around 44\% with \textit{sentence-BERT}. Despite strong retriever performance in candidate retrieval (see~\autoref{irresults}), LLMs' overall performance remains low.
There is a general under-performance on this track when LLM solutions have been used, and given the low performance, we tested domain-specific LLM i.e. BioMistral-7B~\cite{labrak2024biomistral} and we obtained the following results (* refers to the best model result from~\autoref{llm_ir_results}). NCIT-ORDO 69.04\% $<$ 83.01\%$^*$, OMIM-ORDO 57.84\% $<$ 64.14\%$^*$, SNOMED-FMA 33.98\% $>$ 25.64\%$^*$, SNOMED-NCIT(neoplas) 46.24\% $>$ 49.47\%$^*$, and SNOMED-NCIT(pharm) 62.00\% $<$ 68.02\%$^*$. The low performance on all the tasks even with domain-specific LLM, showed a need for a different approach with LLMs for \textsc{Bio-ML} track.

\section{Conclusion}
The proposed LLMs4OM framework highlights the efficacy of LLMs in OM, specifically in aligning diverse ontologies for knowledge engineering. By rigorously evaluating 20 tasks spanning different domains, our framework shows that LLMs, when combined with retriever models and guided by zero-shot prompting while utilizing $C$, $CP$, and $CC$ representations, can surpass traditional OM systems in complex matching scenarios. These findings underscore the significant potential of LLMs in OM, paving the way for further exploration.

\vspace{-2mm}

%
%
\bibliographystyle{splncs04}
\bibliography{refrences}

\end{document}